# A FAST SWITCHING FILTER FOR IMPULSIVE NOISE REMOVAL FROM COLOR IMAGES

**M. Emre Celebi*[1], Hassan A. Kingravi[1], Bakhtiyar Uddin[2], Y. Alp Aslandogan[1]**

[1]Department of Computer Science and Engineering, University of Texas at Arlington, Arlington, TX, USA

[2]T-Mobile, Seattle, WA, USA

## ABSTRACT

In this paper, we present a fast switching filter for impulsive noise removal from color images. The filter exploits the *HSL* color space, and is based on the *peer group* concept, which allows for the fast detection of noise in a neighborhood without resorting to pairwise distance computations between each pixel. Experiments on large set of diverse images demonstrate that the proposed approach is not only extremely fast, but also gives excellent results in comparison to various state-of-the-art filters.

## 1. Introduction

The growing use of color images in diverse applications such as content-based image retrieval, medical image analysis, biometrics, remote sensing, watermarking, and visual quality inspection has led to an increasing interest in color image processing. These applications need to perform many of the same tasks as their grayscale counterparts, such as edge detection, segmentation and feature extraction [1]. However, images are often contaminated with noise which is often introduced during acquisition or transmission. In particular, the introduction of bit errors and impulsive noise into an image not only lowers its perceptual quality but also makes subsequent tasks such as edge detection and segmentation more difficult. Therefore, the removal of noise from an image is often a necessary preprocessing step for these tasks. Modern image filtering solutions can eliminate noise without significantly degrading the underlying image structures such as edges and fine details [2]. Recent applications of color image denoising include enhancement of cDNA microarray images [3,4], virtual restoration of artworks [5,6], and video filtering [7-10].

* Corresponding author (Email: celebi@cse.uta.edu , Phone: 1-682-433-6264)

Numerous filters have been proposed in the literature for noise removal from color images [11-14]. Among these, nonlinear vector filters have proved successful in dealing with impulsive noise while preserving edges and image details [13]. These filters treat pixels in a color image as vectors to avoid color shifts and artifacts. An important class of nonlinear vector filters is the one based on robust order-statistics with the vector median filter (*VMF*) [15], the basic vector directional filter (*BVDF*) [16], and the directional-distance filter (*DDF*) [17] being the most well-known examples. These filters involve the reduced ordering [18] of a set of input vectors within a window to compute the output vector.

The fundamental order-statistics based filters (*VMF*, *BVDF*, and *DDF*) as well as their fuzzy [19,20] and hybrid [21] extensions share a common deficiency in that they are implemented uniformly across the image and tend to modify pixels that are not corrupted by noise [22]. This results in excessive smoothing and the consequent blur of edges and loss of fine image details. In order to overcome this, intelligent filters that switch between a robust order-statistics based filter such as the *VMF* and the identity operation have been introduced [22-37]. These filters determine whether the pixel under consideration is noisy or not in the context of its neighborhood. In the former case, the pixel is replaced by the output of the noise removal filter; otherwise, it is left unchanged to preserve the desired (noise-free) signal structures. Such an approach is computationally efficient considering that the expensive filtering operation is performed only on the noisy pixels, which often comprise a small percentage of the image.

In this paper, we introduce a new switching filter for the removal of impulsive noise from color images. The proposed filter exploits the *HSL* color space [13], and is based on the concept of a *peer group* [22], which allows for the fast detection of noise in a neighborhood without resorting to pairwise distance computations between each pixel. The center pixel in a neighborhood is considered as noise-free if it has a certain number of pixels that are similar to it. In this case, it remains intact. Otherwise, it is replaced by the *VMF* output, i.e. the pixel that minimizes the sum of distances to all other pixels in the neighborhood. The method is tested on a large set of images from diverse domains. The results demonstrate that the proposed filter is not only extremely fast, but also gives excellent results in comparison to various state-of-the-art filters.

The rest of the paper is organized as follows. Section 2 describes the proposed method and the motivation behind the choice of the *HSL* space for the calculation of the similarity metric. Section 3 presents the experimental results. Finally, section 4 gives the conclusions.

## 2. Proposed Method

Let $y(x)$: $Z^2 \rightarrow Z^3$ denote an *RGB* color image that is comprised of a two-dimensional array of three component samples. Although natural images are often nonstationary, filters operate on the assumption that they can be subdivided into small regions that are stationary [12]. This is accomplished using a small window that slides through the individual image pixels while performing the filtering operation locally. The most commonly used window is a square-shaped window $W = \{ x_i \mid i = 1, 2, \ldots, n \}$ of a finite size $n$, where $x_1, x_2, \ldots, x_n$ is a set of pixels centered around $x_{(n+1)/2}$ which determines the position of the window.

Most vector filters operate by ordering the vectors inside the filter window. However, calculating the aggregate distances used in the ordering criterion may limit the use of these filters in real-time applications. One way to reduce the computational requirements of a nonlinear vector filter is to limit the number of comparisons that are performed between the center pixel and the neighboring pixels in the window. The Fast Peer Group Filter (*FPGF*) [31] uses the concept of the *peer group* [22] to determine the output vector according to the following rule:

$$x_{FPGF} = \begin{cases} x_{(n+1)/2} & if \ \left| \left\{ x_j \in W \mid j \neq (n+1)/2 \ and \ \left\| x_{(n+1)/2} - x_j \right\|_p \leq Tol \right\} \right| \geq m \\ x_{VMF} & otherwise \end{cases} \quad (1)$$

where *Tol* is the distance threshold, *m* is the size of the *peer group*, $|.|$ is the set cardinality, $\|.\|_p$ is the $L_p$ (Minkowski) norm, and $x_{VMF}$ denotes the *VMF* output given by:

$$x_{VMF} = \underset{x_i \in W}{argmin} \sum_{j=1}^{n} \left\| x_i - x_j \right\|_p \quad (2)$$

Essentially, the *peer group* of a pixel represents the neighboring pixels in the window that are sufficiently 'similar' to it according to a particular measure. This concept is due to Lee [38] and has been used extensively in the design of various filters, often under the name of *extended spatial neighborhood* [31].

The *FPGF* is much faster than the well-known vector filters mentioned in the previous section because it declares the center pixel to be noise-free as soon as *m* pixels in the window are determined to be sufficiently similar to it. If *m* is low, and the level of noise in the image is not very high, this allows for a dramatic reduction in the number of distance computations that need to be performed. In particular, the minimum and maximum number of distance calculations necessary to classify a pixel equal *m* and *n*-*m*, respectively. Therefore, on the average, the number of distance calculations performed by the *FPGF* is much lower than that performed by the *VMF*, i.e. $n(n-1)/2$. However, due to the nature of the $L_2$ norm, the distance computations performed in highly correlated spaces such as *RGB* remain expensive. On the other hand, if the image is transformed into a color space which decouples chromaticity and luminance, the distance between two color vectors can be evaluated without such a computation. In this study, we adopted the *HSL* color space in order to accomplish this.

The *HSL* (hue, saturation and lightness) color space is an intuitive alternative to the *RGB* space [13]. It uses approximately cylindrical coordinates, and is a non-linear deformation of the *RGB* color cube (Figure 1a). The hue $H \in [0,360]$ is a function of the angle in the polar coordinate system and describes a pure color. The saturation $S \in [0,100]$ is proportional to radial distance and denotes the purity of a color. Finally, the lightness $L \in [0,255]$ is the distance along the axis perpendicular to the polar coordinate plane and represents the brightness. The distance between two vectors $x_i = (h_i, s_i, l_i)$ and $x_j = (h_j, s_j, l_j)$ in the *HSL* space is given by:

$$D(x_i, x_j) = D_{HSL}(x_i, x_j) = \sqrt{s_i^2 + s_j^2 - 2s_i s_j \cos(h_i - h_j) + (l_i - l_j)^2} \qquad (3)$$

**INSERT FIGURE 1 HERE**

Building upon the idea of the *peer group* in much the same way as the *FPGF*, we propose a new filtering algorithm called the Fast *HSL*-based Switching Filter (*FHSF*). First, the *RGB* image is transformed to the *HSL* space [13]. The output vector in a window is then determined according to the following rule:

$$x_{FHSF} = \begin{cases} x_{(n+1)/2} & \text{if } \left|\{x_j \in W \mid j \neq (n+1)/2 \text{ and } S(x_{(n+1)/2}, x_j) = 1\}\right| \geq m \\ x_{VMF} & \text{otherwise} \end{cases}$$
$$S(x_i, x_j) = \begin{cases} 1 & \text{if } |h_i - h_j| \leq Ht \text{ and } |s_i - s_j| \leq St \text{ and } |l_i - l_j| \leq Lt \\ 0 & \text{otherwise} \end{cases} \qquad (4)$$

where $(h_i, s_i, l_i)$ and $(h_j, s_j, l_j)$ denote the hue, saturation, and lightness of the pixels $x_i$ and $x_j$, respectively. *Ht*, *St*, and *Lt* are the thresholds for the hue, saturation, and lightness, respectively.

The *FHSF* algorithm works as follows. First, it checks whether the center pixel is noisy or not. If the pixel is determined to be noisy, it is replaced by the *VMF* output. Otherwise, it remains untouched. A noise-free pixel is one which has a minimum of *m* peers that are sufficiently similar to it. The similarity is determined by the function *S*, which checks to see if the hue, saturation and lightness of the pixel are close to those of the center pixel.

The similarity function *S* is clearly cheaper to evaluate when compared to the $L_2$ norm in the *RGB* space. The superficial similarity between the *S* function and the $L_1$ norm can be discounted by the fact that the former operates in the decorrelated *HSL* space as opposed to the correlated *RGB* space, and consequently the conjunction involved in this function allows for *short-circuit* evaluation. That is, for instance, as long as two color vectors differ in hue, the remaining two conditions need not be evaluated. On the other hand, in the $L_1$ norm the absolute differences between the *R*, *G*, and *B* components always need to be calculated. Table 1 shows the number of elementary operations required by each function. It can be seen that in the worst case, since *COMPs* and *ADDs* have the same complexity [39], the *S* function has the same number of operations as the $L_1$ norm.

**INSERT TABLE 1 HERE**

## 3. Experimental Results

### 3.1 Noise Model and Error Metrics

Several simplified color image noise models have been proposed in the literature [10,11,13]. In this study, the correlated impulsive noise model originally proposed in [10] is adopted. In order to evaluate the filtering performance the following error metrics are used: *MAE* (Mean Absolute Error) [13], *MSE* (Mean Squared Error) [13], *NCD* (Normalized Color Distance) [13], and *PCD* (Perceptual Color Distance) [40-42]. *MAE* and *MSE* are based on the *RGB* color difference and measure the detail preservation and noise suppression capability of a filter, respectively. *NCD* and *PCD* are perceptually oriented metrics that measure the color preservation capability of a filter. *NCD* is based on the *CIELAB* color difference whereas *PCD* is based on the *S-CIELAB* color difference, which is a spatial extension of the former [43]. It should be noted that, to the best of the authors' knowledge, *PCD* has not been used in the color image filtering literature to date. It is included because it takes into account both the spatial and color sensitivity of the human visual system [41].

### 3.2 Parameter Selection

There are four parameters involved in the proposed filter: *m* (the peer group size), *Ht* (Hue Threshold), *St* (Saturation Threshold) and *Lt* (Lightness Threshold). Appropriate ranges for these parameters need to be determined to ensure a good filtering performance on a variety of images. Since the filtering operation is very fast, a simple grid search procedure can be used for this task. In order to do this, the parameter space should first be quantized.

The parameters *m*, *Ht*, *St*, and *Lt* were restricted to [1,8][1] (step size $\Delta = 1$), [6,20] ($\Delta = 2$), [4,16] ($\Delta = 2$), and [32,64] ($\Delta = 4$), respectively. The sizes of the intervals for the *Ht*, *St*, and *Lt* parameters follow the relative importance of the individual components of the *HSL* space. This is because the human visual system is most sensitive to changes in hue, followed by saturation, and then lightness [44]. For example, the hue threshold *Ht* is restricted to the [6,20] interval because for noise removal purposes, two colors that have more than $20^{\circ}$ of hue difference can safely be considered as dissimilar (see Figure 1b).

A set of 100 images was collected from the World Wide Web to be used in the grid search. These included images of people, animals, plants, buildings, aerial maps, man-made objects, natural scenery, paintings, sketches, as well scientific, biomedical, and synthetic images and test images commonly used in the color image processing literature. Figure 2 shows several representative images from this set.

**INSERT FIGURE 2 HERE**

The *PCD* measure was used to quantify the goodness of a particular set of parameters $\{m, Ht, St, Lt\}$. Figure 3 shows the minimum *PCD* values obtained during the grid search at each *m* value for several images that are contaminated with 5%, 10%, 15% and 20% impulsive noise.

**INSERT FIGURE 3 HERE**

As explained in Section 2, the filtering operation is faster for lower values of *m*. In fact, the performance of the proposed filter (in terms of both the effectiveness and the efficiency) will approach that of the *VMF* at high values of $m$. It can be seen from Figure 3 that $m = 3$ provides a good compromise between effectiveness and efficiency. This is in line with the observations of Smolka and Chydzinski [31].

[1] Assuming a 3x3 window

The ranges for the remaining three parameters, $Ht$, $St$ and $Lt$, were determined as follows. For each test image, the parameters were varied in the above-mentioned intervals and the corresponding *PCD* values were calculated. Considering the diversity of the images, it is unreasonable to expect the same parameter combination to give the lowest *PCD* value for each image. Therefore, the parameter combinations that achieved the lowest 5% *PCD* values for each image were recorded. It is expected that a parameter combination that will perform well on a variety of images would appear somewhere in these top 5% lists. The intersection of these lists revealed that the following ranges perform well on the test images, $Ht \in [8,12]$, $St \in [8,14]$, and $Lt \in [40,56]$. For comparison with other filters, the following default values are used: $Ht = 10$, $St = 10$, and $Lt = 48$.

Note that the full range of $H$ is [0,360] and thus acceptable values for $Ht$ lie between 2.22% and 3.33% of this range. Similarly, the range of S is [0,100] and values for $St$ lie between 8.00% and 14.00%. Finally, the range of $L$ is [0,255] and values for $Lt$ lie between 15.62% and 21.87%. This is in line with the above-mentioned fact that the human visual system is most sensitive to changes in hue, followed by saturation, and then lightness [44]. Figure 4 shows an example of this phenomenon wherein a zoomed section of the parrots image is corrupted with 10% noise and then filtered using a parameter configuration in which two of the thresholds are fixed while the other one is relaxed. Figure (c) is the filtering result with the default parameters, (d) is with $Ht$ relaxed by 5% ($Ht = 28$), (e) is with $St$ relaxed by 10% ($St = 20$) and (f) is with $Lt$ relaxed by 12.5% ($Lt = 80$). It can be seen that although the change in the hue threshold is the smallest, the degradation in the filtering result is the greatest. On the other hand, the change in the lightness threshold is the largest, but the filtering result is better than those of (d) and (e).

**INSERT FIGURE 4 HERE**

### 3.3 Comparison with State-of-the-Art Filters

The proposed filter is compared with recent switching filters such as the peer group filter (*PGF*) [22], the adaptive vector median filter (*AVMF*) [26], the fast fuzzy noise reduction filter (*FFNRF*) [32], the fast peer group filter (*FPGF*) [31], the vector sigma filters based on the mean and lowest ranked vectors ($SVMF_{mean}$, $SVMF_{rank}$) [33], and their adaptive counterparts ($ASVMF_{mean}$, $ASVMF_{rank}$) [33]. The traditional filters mentioned in Section 1 (*VMF*, *BVDF*, and *DDF*) are also included in this comparison to highlight the merits of the switching technique. Finally, for comparison purposes, the *FHSF* version that uses the *3D* distance function in the *HSL* space ($FHSF_{HSL}$) and the $L_1$ version of the *FPGF* ($FPGF_1$) are also considered in the experiments. In the following discussion, the standard versions of the *FHSF* and the *FPGF* are denoted as $FHSF_S$ (Equation 4) and $FPGF_2$ (Equation 1 with p = 2), respectively.

Figure 5 shows the filtering results for a zoomed section of the cat image. Parts (c) and (d) show the outputs of the non-switching filters, i.e. the *VMF* and the *DDF*. It can be seen that even though these filters suppress the noise very well, this comes at the expense of the blurring of image details, e.g. the whiskers. On the other hand, the switching filters, i.e. the $FPGF_2$, the *FFNRF*, the *PGF*, and the $FHSF_S$ preserve the details satisfactorily. Among these, the $FHSF_S$ strikes the best balance between noise removal and detail preservation.

**INSERT FIGURE 5 HERE**

Figure 6 shows the filtering results for a section of the pig image and the corresponding difference images. In order to obtain the difference images, the pixelwise absolute differences between the original and the filtered images are multiplied by 5 and then negated. As expected, the *VMF* and the *DDF* outputs show significant differences when compared to the original image. In contrast, the switching filters show a clear improvement in restoring the original image. Among these, it can be seen that the *AVMF*, the *PGF*, and the $FHSF_S$ give the best performance.

**INSERT FIGURE 6 HERE**

Tables 2-4 compare the filters using the criteria described in Section 3.1, i.e. *MAE*, *MSE*, *NCD*, *PCD* and the execution time[2] in seconds. It can be seen that the $FHSF_S$ compares favorably with the best filters in terms of filtering effectiveness, as assessed by the first four criteria. The execution time is also a very important factor which determines the practicality of a noise removal filter. From this perspective, due to their high computational requirements, the non-switching filters in general are not appropriate for denoising large images that are common in domains such as astronomy, remote sensing and biology. Regarding the remaining filters, as the image size increases, the computational advantage of the $FHSF_S$ over the others becomes apparent. In general, the $FHSF_S$ is almost twice as fast as the next fastest filter, i.e. the $FPGF_1$. Note that the timing for the $FHSF_S$ includes the *RGB* to *HSL* transform, although this is negligible[3].

**INSERT TABLE 2 HERE**

**INSERT TABLE 3 HERE**

**INSERT TABLE 4 HERE**

In summary, the experiments demonstrate that the $FHSF_S$ combines simplicity, excellent filtering performance and significant computational efficiency, which makes it a practical method for impulsive noise removal from color images.

## 4. Conclusions

In this paper, we introduced a fast switching filter for the removal of impulsive noise from color images. The proposed filter exploited the *HSL* color space in conjunction with the concept of a *peer group* in order to allow for the fast detection of noise in a neighborhood. The method was tested on a large set of images from diverse domains, as well as classical images used in the color image processing literature. The experiments demonstrated that the new method is much faster than state-of-the-art filters and that the filtering quality is also excellent.

---

[2] C language, GCC 3.4.4 compiler, Intel Centrino 1.6GHz processor

[3] On a 4096*x*4096 image that contains 16 million unique colors, the *RGB* to *HSL* transform takes 45 nanoseconds per pixel. On a typical 1024*x*1024 image, the total transformation time is less than 0.05 seconds.

## Acknowledgements

This work was supported by grants from NSF (#0216500-EIA), Texas Workforce Commission (#3204600182), and James A. Schlipmann Melanoma Cancer Foundation.

## Figure Legend

**Figure 1.** (a) *HSL* double hexcone (b) Hue circle

**Figure 2.** Representative images from the image set

**Figure 3.** *m* vs. minimum *PCD* at noise levels (a) 5%, (b) 10%, (c) 15% and (d) 20%

**Figure 4.** Filtering results for the parrots image using different parameter configurations

**Figure 5.** Filtering results for the cat image corrupted with 10% noise

**Figure 6.** Filtering results for the pig image corrupted with 10% noise and the corresponding absolute difference images

## Table Legend

**Table 1.** Number of elementary operations

**Table 2.** Comparison of the filters on the test images at 5% noise level

**Table 3.** Comparison of the filters on the test images at 10% noise level

**Table 4.** Comparison of the filters on the test images at 15% noise level

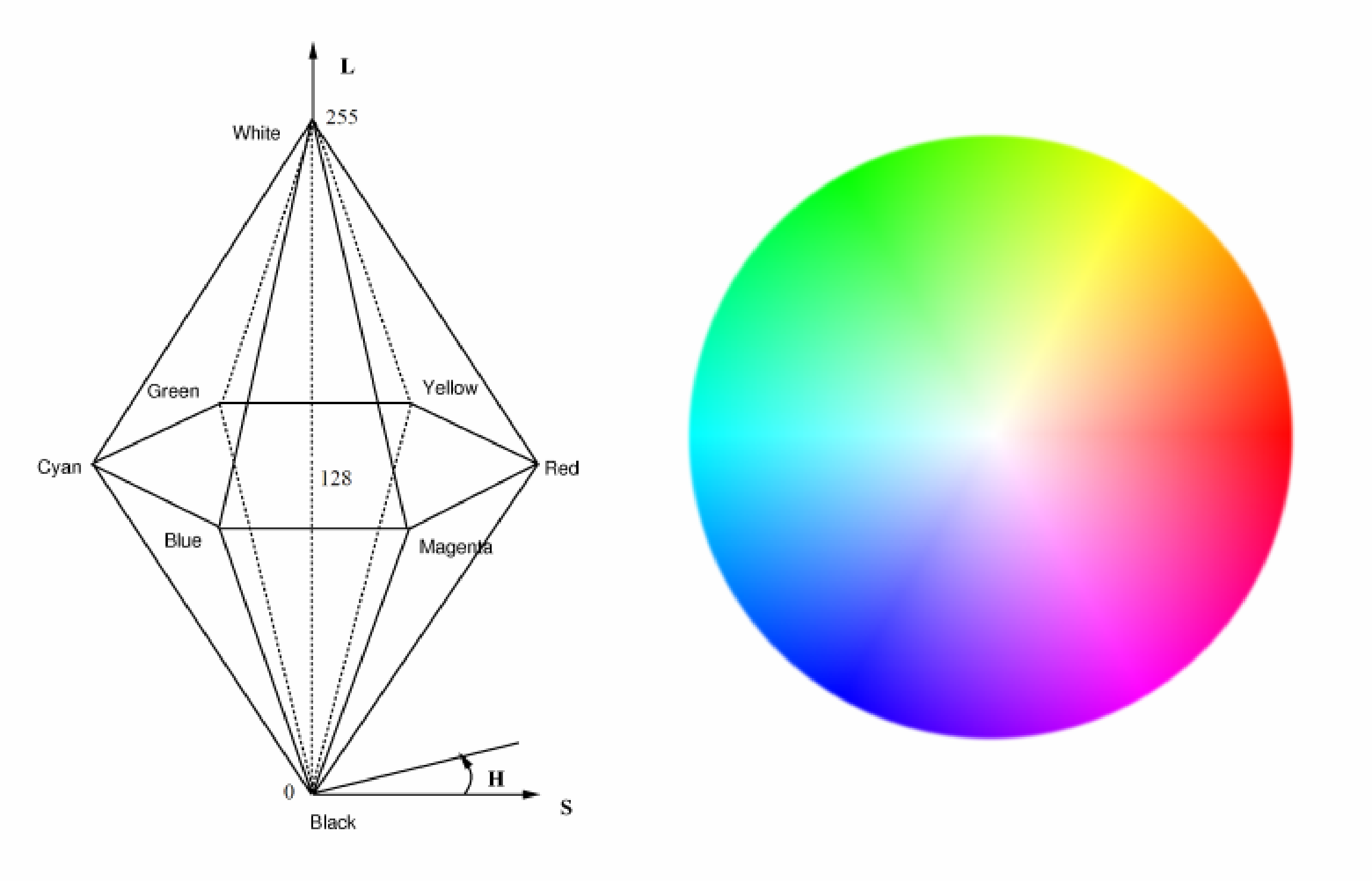

**Figure 1.** (a) *HSL* double hexcone (b) Hue circle

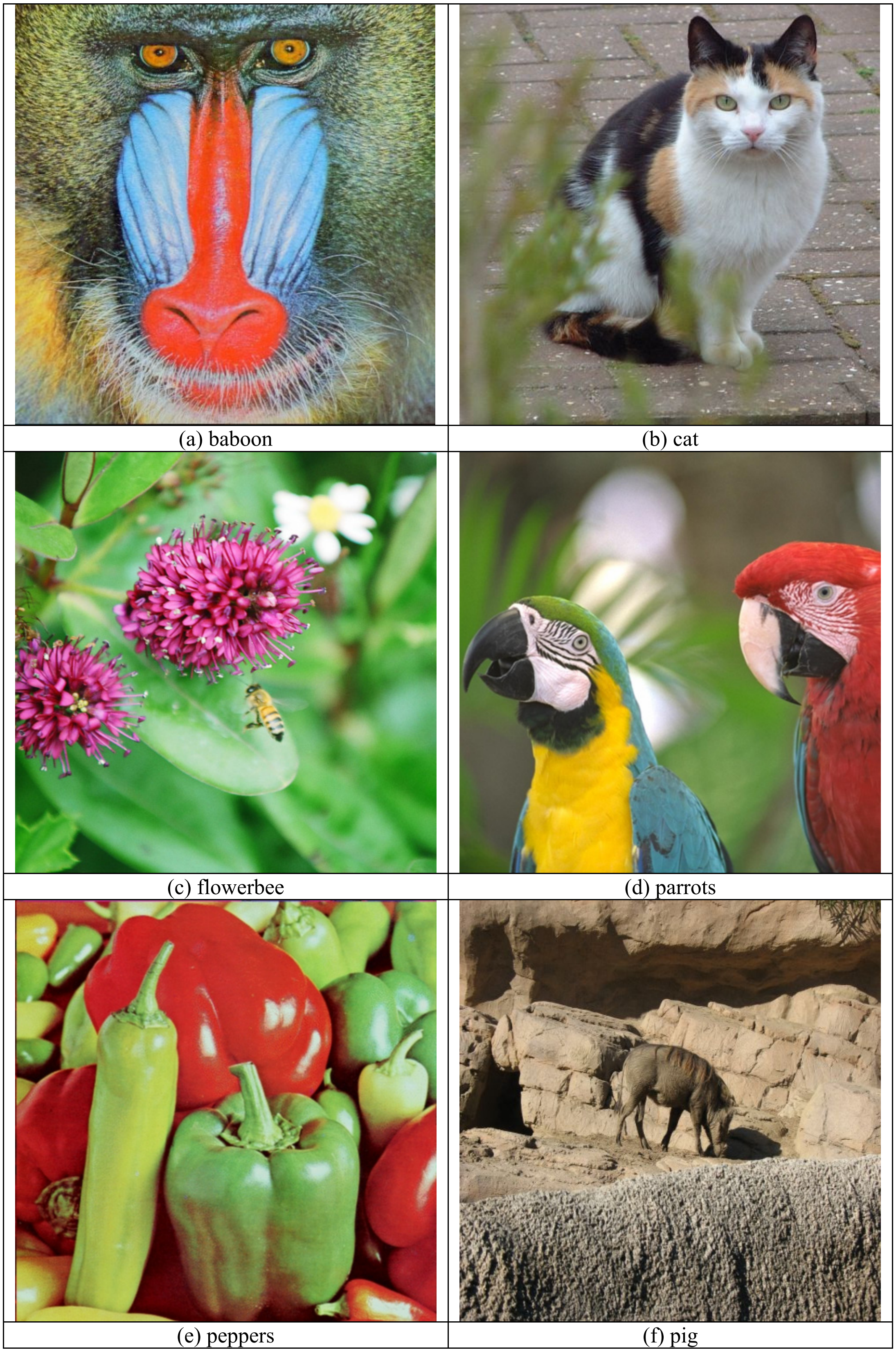

(a) baboon

(b) cat

(c) flowerbee

(d) parrots

(e) peppers

(f) pig

**Figure 2.** Representative images from the image set

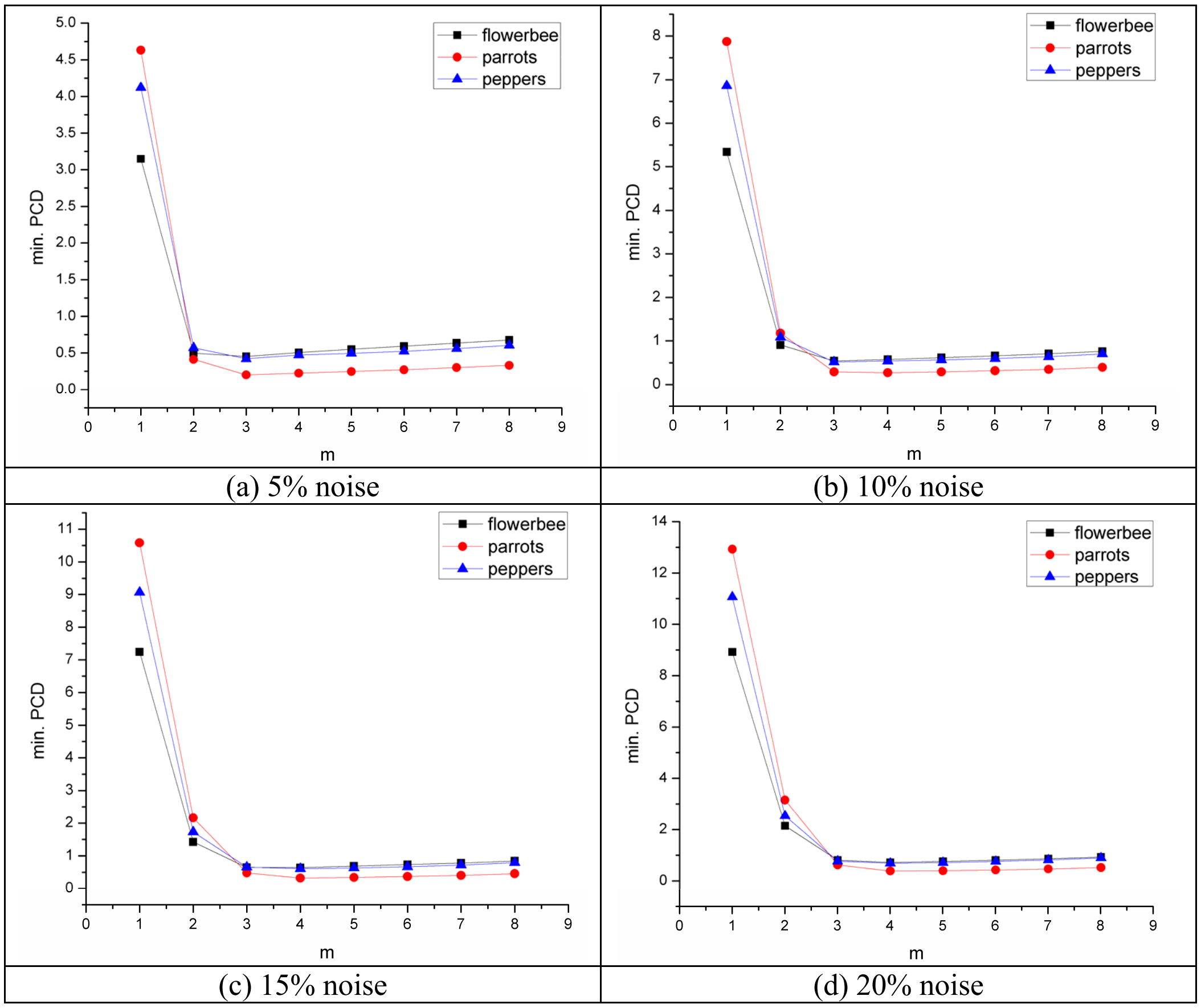

**Figure 3.** *m* vs. minimum *PCD* at noise levels (a) 5%, (b) 10%, (c) 15% and (d) 20%

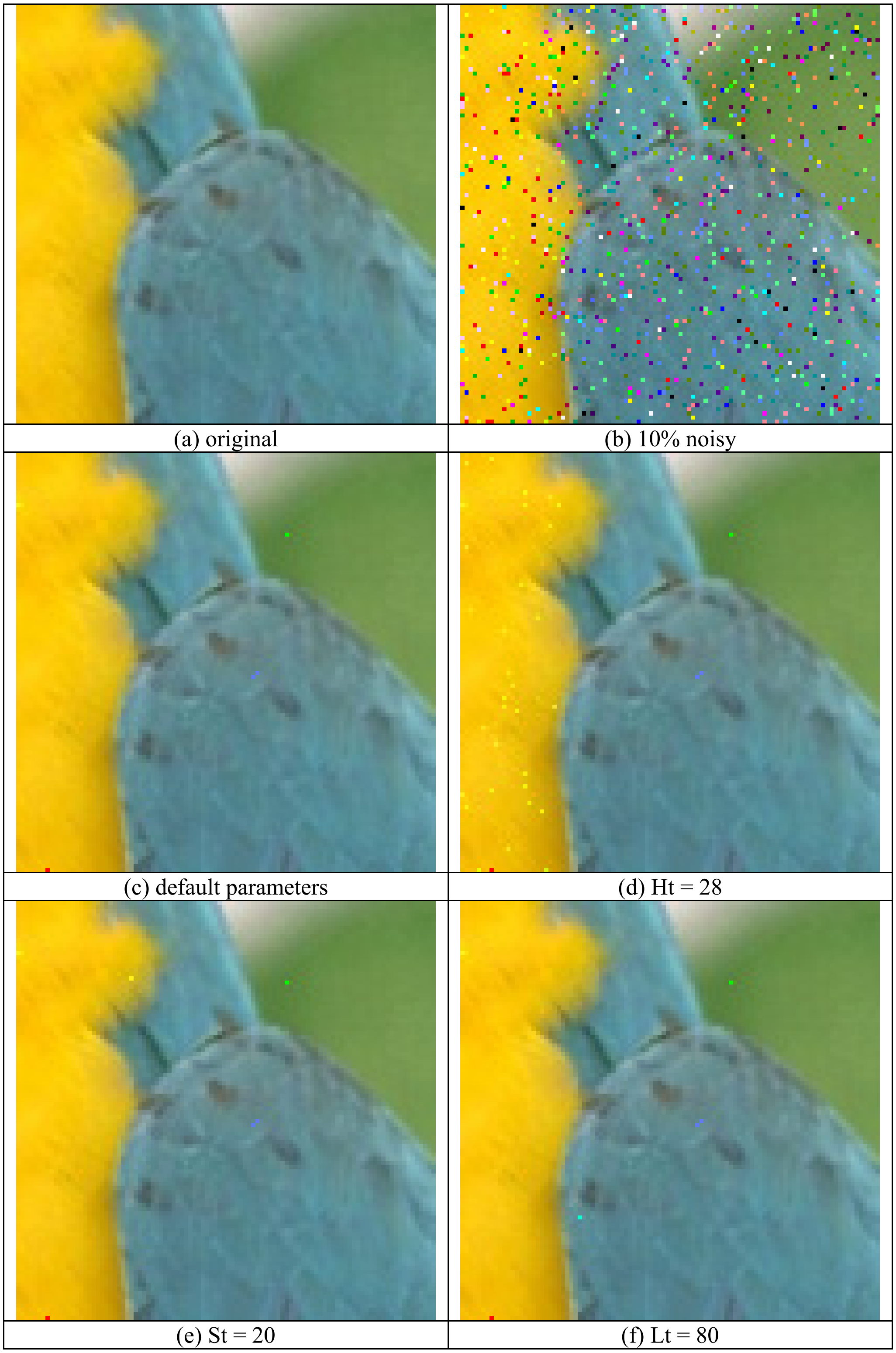

**Figure 4.** Filtering results for the parrots image using different parameter configurations

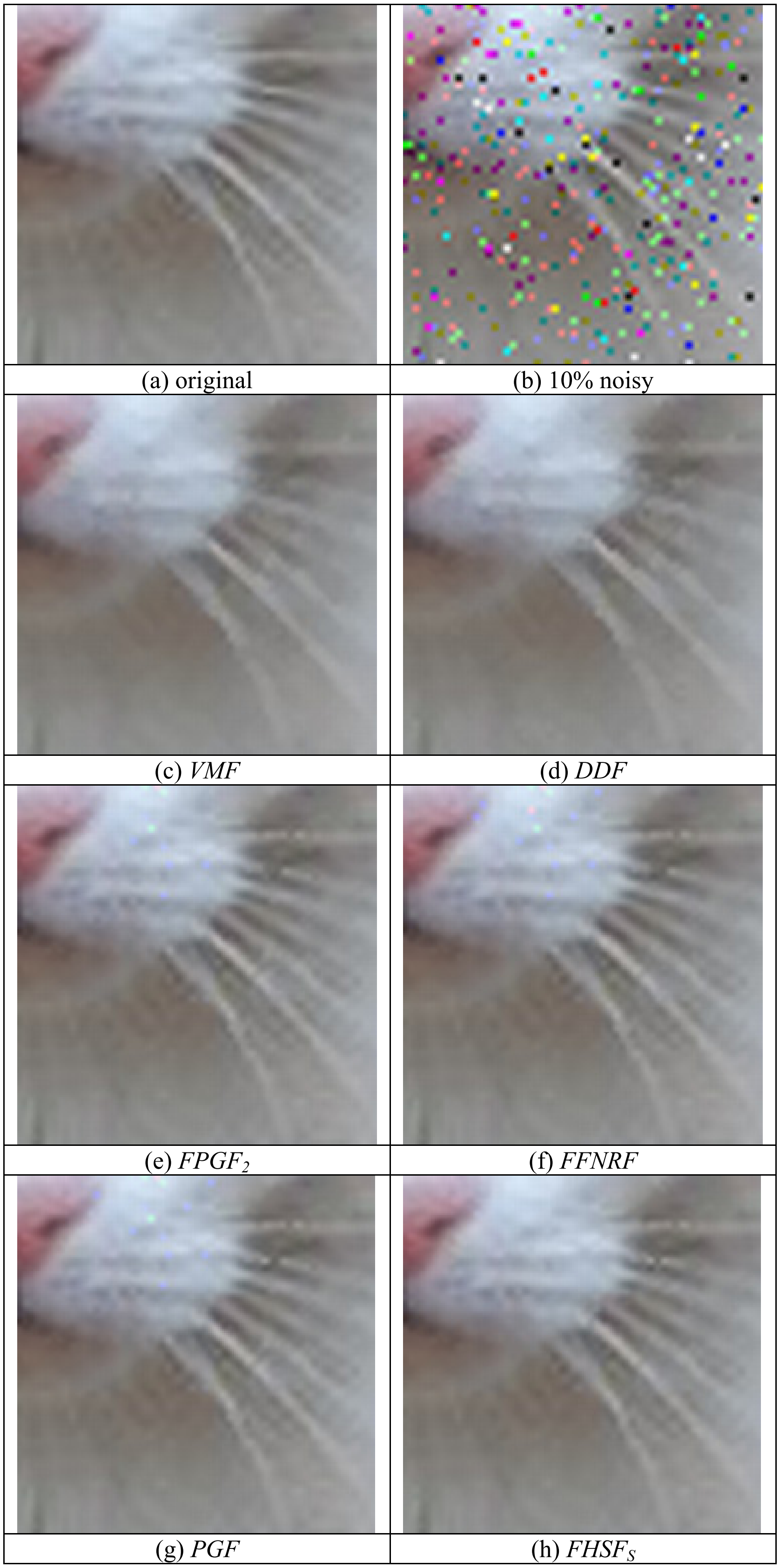

(a) original | (b) 10% noisy | (c) *VMF* | (d) *DDF* | (e) $FPGF_2$ | (f) *FFNRF* | (g) *PGF* | (h) $FHSF_S$

**Figure 5.** Filtering results for the cat image corrupted with 10% noise

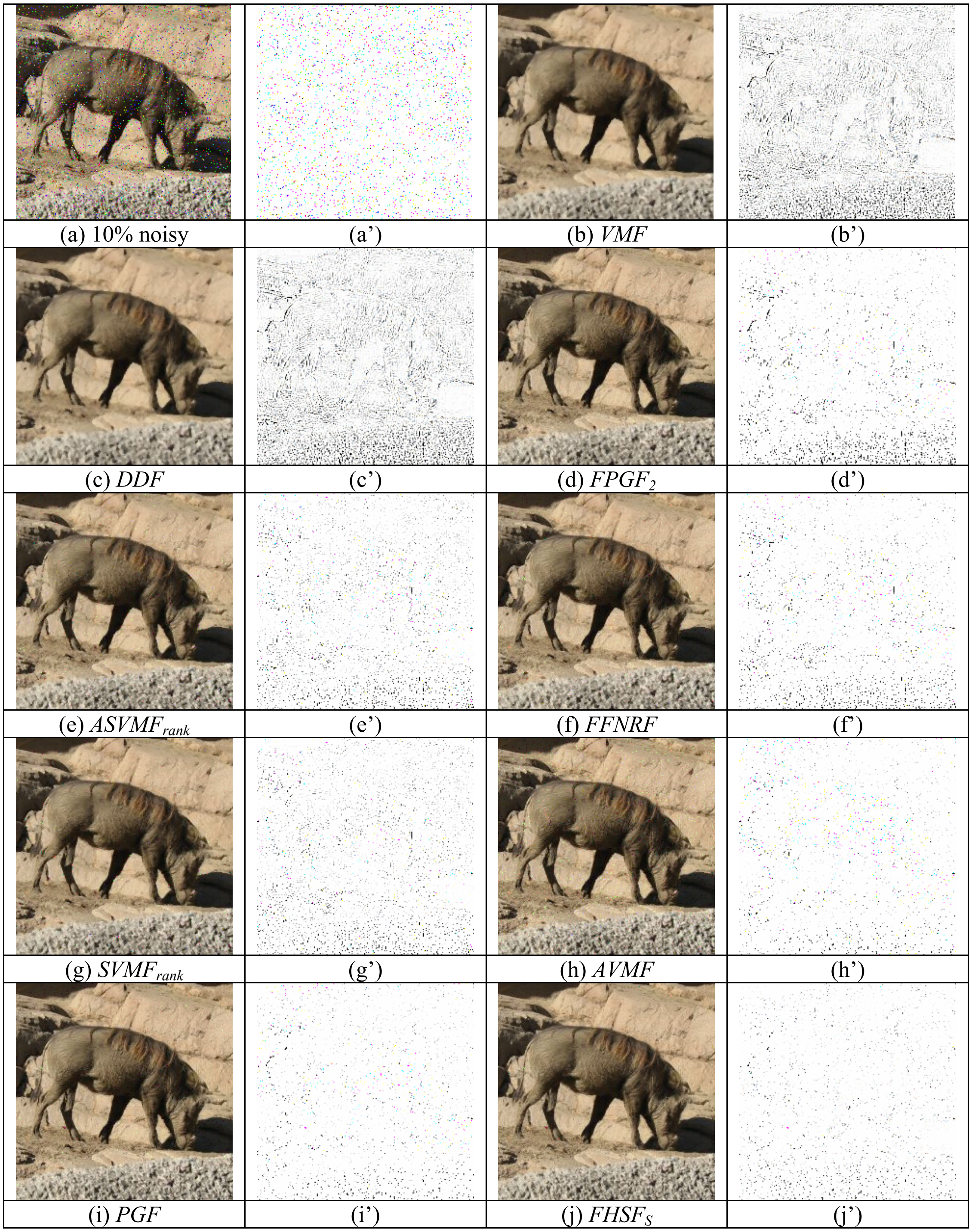

**Figure 6.** Filtering results for the pig image corrupted with 10% noise and the corresponding absolute difference images

**Table 1.** Number of elementary operations

| *Function* | *ABS* | *ADD* | *SUB* | *COMP* | *MULT* | *COS* |
|---|---|---|---|---|---|---|
| $L_1$ | 3 | 2 | 3 | 1 | - | - |
| $L_2$ | - | 2 | 3 | 1 | 3 | - |
| $D_{HSL}$ | - | 2 | 3 | 1 | 6 | 1 |
| $S$ | max. 3 | - | max. 3 | max. 3 | - | - |

**Table 2.** Comparison of the filters on the test images at 5% noise level

| BABOON (512 x 512 pixels) | | | | | |
|---|---|---|---|---|---|
| Filter | MAE | MSE | NCD | PCD | Time |
| NONE | 3.021 | 444.912 | 0.054147 | 3.190 | - |
| ASVMF$_{mean}$ | 5.288 | 213.915 | 0.034924 | 2.245 | 0.265 |
| ASVMF$_{rank}$ | 4.752 | 203.662 | 0.031785 | 2.189 | 0.672 |
| AVMF | 1.909 | 114.535 | 0.017017 | 1.263 | 0.828 |
| BVDF | 11.270 | 379.708 | 0.07534 | 3.363 | 8.281 |
| DDF | 10.293 | 315.996 | 0.068711 | 3.069 | 8.765 |
| FFNRF | 4.044 | 218.383 | 0.027072 | 1.984 | 0.375 |
| FHSF$_{HSL}$ | 5.120 | 202.901 | 0.034014 | 2.132 | 0.359 |
| **FHSF$_{S}$** | **2.443** | **102.198** | **0.016858** | **1.269** | **0.093** |
| FPGF$_{2}$ | 4.416 | 201.745 | 0.028855 | 2.060 | 0.234 |
| FPGF$_{1}$ | 7.164 | 271.991 | 0.046831 | 2.660 | 0.266 |
| PGF | 1.483 | 69.330 | 0.010498 | 0.998 | 0.250 |
| SVMF$_{mean}$ | 4.015 | 169.237 | 0.026675 | 1.930 | 0.359 |
| SVMF$_{rank}$ | 4.010 | 169.825 | 0.026523 | 1.927 | 0.594 |
| VMF | 10.570 | 316.689 | 0.071926 | 3.171 | 0.624 |

| PEPPERS (512 x 480) | | | | | |
|---|---|---|---|---|---|
| Filter | MAE | MSE | NCD | PCD | Time |
| NONE | 3.068 | 489.064 | 0.047504 | 4.138 | - |
| ASVMF$_{mean}$ | 0.506 | 6.389 | 0.004151 | 0.534 | 0.172 |
| ASVMF$_{rank}$ | 0.507 | 7.089 | 0.004236 | 0.542 | 0.641 |
| AVMF | 0.419 | 21.954 | 0.006366 | 0.613 | 0.750 |
| BVDF | 2.150 | 30.059 | 0.018474 | 1.254 | 7.704 |
| DDF | 1.730 | 15.445 | 0.014379 | 0.922 | 8.047 |
| FFNRF | 0.212 | 4.908 | 0.002637 | 0.436 | 0.329 |
| FHSF$_{HSL}$ | 0.233 | 3.021 | 0.002206 | 0.441 | 0.235 |
| **FHSF$_{S}$** | **0.208** | **2.672** | **0.002091** | **0.430** | **0.078** |
| FPGF$_{2}$ | 0.220 | 3.885 | 0.002388 | 0.440 | 0.125 |
| FPGF$_{1}$ | 0.266 | 4.260 | 0.002657 | 0.471 | 0.109 |
| PGF | 0.207 | 4.337 | 0.002422 | 0.431 | 0.234 |
| SVMF$_{mean}$ | 0.380 | 4.911 | 0.003151 | 0.489 | 0.312 |
| SVMF$_{rank}$ | 0.335 | 3.642 | 0.00265 | 0.475 | 0.562 |
| VMF | 1.680 | 10.600 | 0.014163 | 0.866 | 0.563 |

| PARROTS (1536 x 1024) | | | | | |
|---|---|---|---|---|---|
| Filter | MAE | MSE | NCD | PCD | Time |
| NONE | 3.065 | 472.017 | 0.061343 | 4.685 | - |
| ASVMF$_{mean}$ | 0.179 | 3.110 | 0.002168 | 0.1465 | 1.187 |
| ASVMF$_{rank}$ | 0.181 | 3.653 | 0.002350 | 0.147 | 3.547 |
| AVMF | 0.359 | 22.315 | 0.007956 | 0.251 | 4.390 |
| BVDF | 0.861 | 8.135 | 0.007753 | 0.384 | 39.719 |
| DDF | 0.583 | 3.936 | 0.005536 | 0.290 | 43.391 |
| FFNRF | 0.101 | 2.449 | 0.001768 | 0.107 | 1.906 |
| FHSF$_{HSL}$ | 0.082 | 1.220 | 0.000946 | 0.108 | 1.204 |
| **FHSF$_{S}$** | **0.065** | **0.855** | **0.000741** | **0.097** | **0.407** |
| FPGF$_{2}$ | 0.107 | 2.263 | 0.001714 | 0.103 | 0.735 |
| FPGF$_{1}$ | 0.125 | 2.332 | 0.001708 | 0.111 | 0.547 |
| PGF | 0.104 | 2.608 | 0.001832 | 0.106 | 1.485 |
| SVMF$_{mean}$ | 0.123 | 2.042 | 0.001472 | 0.120 | 1.922 |
| SVMF$_{rank}$ | 0.105 | 1.178 | 0.001072 | 0.111 | 3.500 |
| VMF | 0.540 | 2.609 | 0.005351 | 0.267 | 3.500 |

| FLOWERBEE (3088 x 2048) | | | | | |
|---|---|---|---|---|---|
| Filter | MAE | MSE | NCD | PCD | Time |
| NONE | 3.066 | 480.868 | 0.046835 | 1.717 | - |
| ASVMF$_{mean}$ | 0.578 | 5.952 | 0.003608 | 0.272 | 5.094 |
| ASVMF$_{rank}$ | 0.559 | 6.543 | 0.003593 | 0.278 | 14.687 |
| AVMF | 0.376 | 20.639 | 0.005393 | 0.354 | 17.359 |
| BVDF | 1.814 | 11.650 | 0.010962 | 0.426 | 184.969 |
| DDF | 1.655 | 9.473 | 0.009879 | 0.394 | 201.374 |
| FFNRF | 0.167 | 2.630 | 0.001547 | 0.174 | 7.891 |
| FHSF$_{HSL}$ | 0.175 | 1.718 | 0.001219 | 0.173 | 4.907 |
| **FHSF$_{S}$** | **0.144** | **1.313** | **0.00107** | **0.163** | **1.453** |
| FPGF$_{2}$ | 0.160 | 2.149 | 0.001369 | 0.166 | 2.969 |
| FPGF$_{1}$ | 0.180 | 2.168 | 0.001383 | 0.173 | 2.281 |
| PGF | 0.167 | 2.807 | 0.001518 | 0.171 | 5.969 |
| SVMF$_{mean}$ | 0.439 | 4.209 | 0.002671 | 0.243 | 8.344 |
| SVMF$_{rank}$ | 0.417 | 3.390 | 0.002406 | 0.235 | 14.047 |
| VMF | 1.697 | 9.660 | 0.010444 | 0.397 | 14.094 |

**Table 3.** Comparison of the filters on the test images at 10% noise level

| BABOON (512 x 512 pixels) | | | | | |
|---|---|---|---|---|---|
| Filter | MAE | MSE | NCD | PCD | Time |
| NONE | 6.168 | 914.459 | 0.109563 | 5.505 | - |
| $ASVMF_{mean}$ | 5.014 | 210.446 | 0.035257 | 2.233 | 0.281 |
| $ASVMF_{rank}$ | 4.619 | 205.536 | 0.033701 | 2.196 | 0.657 |
| AVMF | 2.680 | 149.962 | 0.026662 | 1.644 | 0.797 |
| BVDF | 11.650 | 397.573 | 0.078111 | 3.508 | 8.172 |
| DDF | 10.564 | 324.853 | 0.070965 | 3.153 | 8.813 |
| FFNRF | 4.485 | 231.183 | 0.031609 | 2.161 | 0.375 |
| $FHSF_{HSL}$ | 5.768 | 222.239 | 0.038852 | 2.333 | 0.375 |
| **$FHSF_{S}$** | **3.151** | **127.181** | **0.022178** | **1.539** | **0.109** |
| $FPGF_{2}$ | 5.205 | 224.401 | 0.034677 | 2.310 | 0.235 |
| $FPGF_{1}$ | 7.770 | 287.367 | 0.051368 | 2.819 | 0.281 |
| PGF | 2.288 | 98.839 | 0.016958 | 1.356 | 0.329 |
| $SVMF_{mean}$ | 4.006 | 172.698 | 0.028911 | 1.978 | 0.422 |
| $SVMF_{rank}$ | 4.041 | 173.619 | 0.028697 | 1.984 | 0.656 |
| VMF | 10.813 | 326.192 | 0.073878 | 3.256 | 0.641 |

| PEPPERS (512 x 480) | | | | | |
|---|---|---|---|---|---|
| Filter | MAE | MSE | NCD | PCD | Time |
| NONE | 6.184 | 983.288 | 0.094969 | 6.764 | - |
| $ASVMF_{mean}$ | 0.646 | 15.608 | 0.006727 | 0.618 | 0.203 |
| $ASVMF_{rank}$ | 0.679 | 18.524 | 0.007333 | 0.646 | 0.578 |
| AVMF | 0.845 | 44.461 | 0.01275 | 0.851 | 0.703 |
| BVDF | 2.378 | 40.246 | 0.021017 | 1.488 | 7.703 |
| DDF | 1.888 | 17.109 | 0.016101 | 0.970 | 7.953 |
| FFNRF | 0.439 | 11.573 | 0.005574 | 0.555 | 0.328 |
| $FHSF_{HSL}$ | 0.3923 | 6.046 | 0.003907 | 0.523 | 0.235 |
| **$FHSF_{S}$** | **0.370** | **6.098** | **0.003869** | **0.508** | **0.079** |
| $FPGF_{2}$ | 0.429 | 7.772 | 0.004726 | 0.527 | 0.157 |
| $FPGF_{1}$ | 0.477 | 7.680 | 0.004843 | 0.555 | 0.140 |
| PGF | 0.432 | 10.999 | 0.005185 | 0.533 | 0.250 |
| $SVMF_{mean}$ | 0.539 | 13.218 | 0.005628 | 0.575 | 0.344 |
| $SVMF_{rank}$ | 0.461 | 8.509 | 0.004435 | 0.529 | 0.563 |
| VMF | 1.842 | 13.163 | 0.015854 | 0.924 | 0.547 |

| PARROTS (1536 x 1024) | | | | | |
|---|---|---|---|---|---|
| Filter | MAE | MSE | NCD | PCD | Time |
| NONE | 6.119 | 941.234 | 0.122676 | 7.956 | - |
| $ASVMF_{mean}$ | 0.310 | 10.874 | 0.005117 | 0.222 | 1.250 |
| $ASVMF_{rank}$ | 0.336 | 12.633 | 0.005785 | 0.221 | 3.610 |
| AVMF | 0.718 | 44.311 | 0.015853 | 0.419 | 4.250 |
| BVDF | 0.935 | 9.100 | 0.008498 | 0.412 | 40.797 |
| DDF | 0.646 | 4.444 | 0.006251 | 0.313 | 43.906 |
| FFNRF | 0.218 | 6.047 | 0.003981 | 0.160 | 1.953 |
| $FHSF_{HSL}$ | 0.149 | 3.201 | 0.001901 | 0.161 | 1.344 |
| **$FHSF_{S}$** | **0.132** | **3.448** | **0.001691** | **0.162** | **0.485** |
| $FPGF_{2}$ | 0.214 | 4.490 | 0.003445 | 0.144 | 0.905 |
| $FPGF_{1}$ | 0.227 | 4.236 | 0.003273 | 0.150 | 0.718 |
| PGF | 0.226 | 6.976 | 0.00403 | 0.177 | 1.531 |
| $SVMF_{mean}$ | 0.238 | 7.857 | 0.003813 | 0.196 | 2.047 |
| $SVMF_{rank}$ | 0.170 | 3.210 | 0.002165 | 0.149 | 3.265 |
| VMF | 0.616 | 3.247 | 0.006174 | 0.296 | 3.250 |

| FLOWERBEE (3088 x 2048) | | | | | |
|---|---|---|---|---|---|
| Filter | MAE | MSE | NCD | PCD | Time |
| NONE | 6.129 | 960.795 | 0.093671 | 2.851 | - |
| $ASVMF_{mean}$ | 0.6449 | 12.948 | 0.005325 | 0.326 | 5.250 |
| $ASVMF_{rank}$ | 0.655 | 15.074 | 0.005739 | 0.346 | 14.797 |
| AVMF | 0.762 | 41.588 | 0.010859 | 0.566 | 17.595 |
| BVDF | 1.925 | 13.044 | 0.01177 | 0.452 | 183.641 |
| DDF | 1.737 | 10.189 | 0.010584 | 0.411 | 192.375 |
| FFNRF | 0.348 | 6.361 | 0.003401 | 0.235 | 8.016 |
| $FHSF_{HSL}$ | 0.307 | 4.138 | 0.002272 | 0.217 | 5.578 |
| **$FHSF_{S}$** | **0.274** | **3.904** | **0.002131** | **0.209** | **1.828** |
| $FPGF_{2}$ | 0.328 | 4.430 | 0.002792 | 0.212 | 3.813 |
| $FPGF_{1}$ | 0.348 | 4.121 | 0.002714 | 0.218 | 2.938 |
| PGF | 0.355 | 7.583 | 0.003343 | 0.233 | 6.625 |
| $SVMF_{mean}$ | 0.524 | 9.415 | 0.004118 | 0.284 | 8.859 |
| $SVMF_{rank}$ | 0.470 | 5.194 | 0.003169 | 0.252 | 14.235 |
| VMF | 1.778 | 10.406 | 0.011118 | 0.417 | 14.156 |

**Table 4.** Comparison of the filters on the test images at 15% noise level

BABOON (512 x 512 pixels)

| Filter | MAE | MSE | NCD | PCD | Time |
|---|---|---|---|---|---|
| NONE | 9.212 | 1357.231 | 0.163820 | 7.247 | - |
| $ASVMF_{mean}$ | 5.043 | 222.431 | 0.039034 | 2.319 | 0.234 |
| $ASVMF_{rank}$ | 4.726 | 219.993 | 0.038661 | 2.285 | 0.610 |
| AVMF | 3.522 | 190.028 | 0.037089 | 1.957 | 0.719 |
| BVDF | 12.032 | 417.613 | 0.080963 | 3.640 | 8.109 |
| DDF | 10.846 | 335.445 | 0.073274 | 3.246 | 8.672 |
| FFNRF | 5.017 | 251.577 | 0.037636 | 2.341 | 0.344 |
| $FHSF_{HSL}$ | 6.483 | 246.382 | 0.044302 | 2.519 | 0.360 |
| **$FHSF_{S}$** | **3.937** | **157.523** | **0.028343** | **1.805** | **0.110** |
| $FPGF_{2}$ | 6.042 | 249.839 | 0.040849 | 2.524 | 0.281 |
| $FPGF_{1}$ | 8.400 | 305.442 | 0.056120 | 2.959 | 0.312 |
| PGF | 3.135 | 132.236 | 0.023939 | 1.679 | 0.328 |
| $SVMF_{mean}$ | 4.205 | 189.528 | 0.03397 | 2.106 | 0.391 |
| $SVMF_{rank}$ | 4.258 | 189.290 | 0.033357 | 2.086 | 0.594 |
| VMF | 11.066 | 337.448 | 0.075857 | 3.343 | 0.594 |

PEPPERS (512 x 480)

| Filter | MAE | MSE | NCD | PCD | Time |
|---|---|---|---|---|---|
| NONE | 9.200 | 1465.480 | 0.141325 | 9.101 | - |
| $ASVMF_{mean}$ | 0.902 | 32.168 | 0.010989 | 0.812 | 0.187 |
| $ASVMF_{rank}$ | 0.962 | 36.363 | 0.012038 | 0.843 | 0.578 |
| AVMF | 1.240 | 64.149 | 0.018588 | 1.097 | 0.687 |
| BVDF | 2.623 | 57.374 | 0.023845 | 1.812 | 7.657 |
| DDF | 2.039 | 19.385 | 0.017644 | 1.043 | 7.953 |
| FFNRF | 0.691 | 20.228 | 0.009018 | 0.709 | 0.328 |
| $FHSF_{HSL}$ | 0.590 | 14.166 | 0.006159 | 0.680 | 0.249 |
| **$FHSF_{S}$** | **0.566** | **14.178** | **0.006134** | **0.676** | **0.094** |
| $FPGF_{2}$ | 0.645 | 11.784 | 0.007078 | 0.632 | 0.172 |
| $FPGF_{1}$ | 0.691 | 11.280 | 0.00703 | 0.663 | 0.156 |
| PGF | 0.677 | 20.420 | 0.008287 | 0.735 | 0.281 |
| $SVMF_{mean}$ | 0.803 | 29.786 | 0.009688 | 0.79 | 0.344 |
| $SVMF_{rank}$ | 0.659 | 17.584 | 0.007142 | 0.668 | 0.547 |
| VMF | 1.987 | 15.428 | 0.017381 | 0.996 | 0.563 |

PARROTS (1536 x 1024)

| Filter | MAE | MSE | NCD | PCD | Time |
|---|---|---|---|---|---|
| NONE | 9.216 | 1419.074 | 0.184092 | 10.680 | - |
| $ASVMF_{mean}$ | 0.536 | 25.797 | 0.010298 | 0.379 | 1.297 |
| $ASVMF_{rank}$ | 0.597 | 29.275 | 0.011644 | 0.368 | 3.500 |
| AVMF | 1.086 | 66.930 | 0.023909 | 0.585 | 4.187 |
| BVDF | 1.017 | 11.178 | 0.009398 | 0.456 | 41.063 |
| DDF | 0.716 | 5.083 | 0.007065 | 0.343 | 44.156 |
| FFNRF | 0.362 | 11.560 | 0.006896 | 0.236 | 1.984 |
| $FHSF_{HSL}$ | 0.237 | 8.304 | 0.003368 | 0.276 | 1.547 |
| **$FHSF_{S}$** | **0.225** | **9.875** | **0.00328** | **0.304** | **0.563** |
| $FPGF_{2}$ | 0.328 | 6.961 | 0.005292 | 0.192 | 1.094 |
| $FPGF_{1}$ | 0.338 | 6.384 | 0.004945 | 0.198 | 0.906 |
| PGF | 0.376 | 14.717 | 0.006867 | 0.311 | 1.703 |
| $SVMF_{mean}$ | 0.449 | 22.600 | 0.008443 | 0.391 | 2.172 |
| $SVMF_{rank}$ | 0.287 | 9.150 | 0.004406 | 0.244 | 3.375 |
| VMF | 0.697 | 4.048 | 0.007065 | 0.331 | 3.360 |

FLOWERBEE (3088 x 2048)

| Filter | MAE | MSE | NCD | PCD | Time |
|---|---|---|---|---|---|
| NONE | 9.179 | 1439.494 | 0.140248 | 3.808 | - |
| $ASVMF_{mean}$ | 0.841 | 27.884 | 0.008753 | 0.441 | 5.312 |
| $ASVMF_{rank}$ | 0.888 | 31.652 | 0.009704 | 0.483 | 14.469 |
| AVMF | 1.148 | 62.520 | 0.016352 | 0.763 | 17.250 |
| BVDF | 2.048 | 15.394 | 0.012673 | 0.484 | 176.156 |
| DDF | 1.825 | 11.110 | 0.011324 | 0.430 | 186.750 |
| FFNRF | 0.551 | 12.041 | 0.005667 | 0.304 | 7.875 |
| $FHSF_{HSL}$ | 0.458 | 9.270 | 0.003634 | 0.270 | 6.250 |
| **$FHSF_{S}$** | **0.427** | **9.834** | **0.003582** | **0.268** | **2.124** |
| $FPGF_{2}$ | 0.502 | 7.020 | 0.004282 | 0.254 | 4.421 |
| $FPGF_{1}$ | 0.523 | 6.352 | 0.00410 | 0.259 | 3.501 |
| PGF | 0.568 | 15.801 | 0.005609 | 0.309 | 7.157 |
| $SVMF_{mean}$ | 0.727 | 23.667 | 0.007241 | 0.384 | 8.907 |
| $SVMF_{rank}$ | 0.600 | 11.385 | 0.004855 | 0.300 | 13.906 |
| VMF | 1.864 | 11.356 | 0.011815 | 0.437 | 13.875 |